\pdfoutput=1

\documentclass{article} 
\usepackage{iclr2016_conference,times}
\usepackage{amsmath}
\usepackage{amssymb}
\usepackage{url}
\usepackage{pgfplots}
\usepackage[export]{adjustbox}

\DeclareMathOperator*{\argmin}{arg\,min}

\title{How (not) to Train your Generative Model: Scheduled Sampling, Likelihood, Adversary?}

\author{Ferenc Husz\'{a}r\\
Balderton Capital LLP, London, UK\\
\texttt{ferenc.huszar@gmail.com}
}

\begin{document}

\maketitle

\begin{abstract}
Modern applications and progress in deep learning research have created renewed interest for generative models of text and of images. However, even today it is unclear what objective functions one should use to train and evaluate these models. In this paper we present two contributions.

Firstly, we present a critique of scheduled sampling, a state-of-the-art training method that contributed to the winning entry to the MSCOCO image captioning benchmark in 2015. Here we show that despite this impressive empirical performance, the objective function underlying scheduled sampling is improper and leads to an inconsistent learning algorithm.

Secondly, we revisit the problems that scheduled sampling was meant to address, and present an alternative interpretation. We argue that maximum likelihood is an inappropriate training objective when the end-goal is to generate natural-looking samples. We go on to derive an ideal objective function to use in this situation instead. We introduce a generalisation of adversarial training, and show how such method can interpolate between maximum likelihood training and our ideal training objective. To our knowledge this is the first theoretical analysis that explains why adversarial training tends to produce samples with higher perceived quality.
\end{abstract}

\section{Introduction}

Building sophisticated generative models that produce realistic-looking images or text is an important current frontier of unsupervised learning. The renewed interest in generative models can be attributed to two factors. Firstly, thanks to the active investment in machine learning by internet companies, we now have several products and practical use-cases for generative models: texture generation\citep{han2008texturesynthesis}, speech synthesis \citep{ou2012speechsynthesis}, image caption generation \citep{lin2014microsoftcoco,vinyals2014captiongeneration}, machine translation \citep{sutskever2014sequencetosequence}, conversation and dialogue generation \citep{vinyals2015conversation,sordoni2015conversation}. Secondly, recent success in generative models, particularly those based on deep representation learning, have raised hopes that our systems may one day reach the sophistication required in these practical use cases. 

While noticable progress has been made in generative modelling, in many applications we are still far from generating fully realistic samples. One of the key open questions is what objective functions one should use to train and evaluate generative models \citep{theis2015generative}. The model likelihood is often considered the most principled training objective and most research in the past decades has focussed on maximum lieklihood(ML) and approximations thereof \cite{hinton2006deepbelief, hyvarinen2006pseudolikelihood, kingma2013variauto}. Recently we have seen promising new training strategies such as those based on adversarial networks \citep{goodfellow2014adversarial, denton2015lapgan} and kernel moment matching \citep{li2015momentmatching,dziugaite2015mmd} which are not | at least on the surface | related to maximum likelihood. Most of this departure from ML was motivated by the fact that the exact likelihood is intractable in the most models. However, some authors have recently observed that even in models whose likelihood is tractable, ML training leads to undesired behaviour, and introduced new training procedures that deliberately differ from maximum likelihood. Here we will focus on scheduled sampling \citep{bengio2015scheduledsampling} which is an example of this.

In this paper we attempt to clarify what objective functions might work well for the generative scenario and which ones should one avoid. In line with \citep{theis2015generative} and \citep{lacoste2011approximate}, we believe that the objective function used for training should reflect the task we want to ultimately use the model for. In the context of this paper, we focus on generative models that are created with the sole purpose of generating realistic-looking samples from. This narrower definition extends to use-cases such as image captioning, texture generation, machine translation and dialogue systems, but excludes tasks such as unsupervised pre-training for supervised learning, semisupervised learning, data compression, denoising and many others.

This paper is organised around the following main contributions:

\begin{description}
\item[scheduled sampling is improper:] In the first half of this paper we focus on autoregressive models for sequence generation. These models are interesting for us mainly because exact maximum likelihood training is tractable, even in relatively complex models such as stacked LSTMs \citep{bengio2015scheduledsampling,sutskever2014sequencetosequence,theis2015lstm}. However, it has been observed that autoregressive generative models trained via ML have some undesired behaviour when they are used to generate samples. We revisit a recent attempt to remedy these problems: scheduled sampling. We reexpress the scheduled sampling training objective in terms of Kullback-Leibler divergences, and show that it is in fact an improper training objective. Therefore we recommend to use scheduled sampling with care.
\item[KL-divergence as a model of perceptual loss:] In the latter part of the paper we seek an alternative solution to the problem scheduled sampling was meant to address. We uncover a more fundamental problem that applies to all generative models: that the likelihood is not the right training objective when the goal is to generate realistic samples. Maximum likelihood can be thought of as minimising the Kullback-Leibler divergence $KL[P\|Q]$ between the real data distribution $P$ and the probabilistic model $Q$. We present a model that suggests generative models should instead be trained to minimise $KL[Q\|P]$, the Kullback-Leibler divergence in the opposite direction. The differences between minimising $KL[P\|Q]$ and $KL[Q\|P]$ are well understood, and explain the observed undesirable behaviour in autoregressive sequence models.

\item[generalised adversarial training:] Unfortunately, $KL[Q\|P]$ is even harder to optimise than the likelihood, so it is unlikely to yield a viable training procedure. Instead, we suggest to minimise an information quantity which we call generalised Jensen-Shannon divergence. We show that this divergence can effectively interpolate between the behaviour of $KL[P\|Q]$ and $KL[P\|Q]$, thereby containing both maximum likelihood, and our ideal perceptual objective function as a special case. We also show that generalisations of the adversarial training procedure proposed in \citep{goodfellow2014adversarial} can be employed to approximately minimise this divergence function. Our analysis also provides a new theoretical explanation for the success of adversarial training in producing qualitatively superior samples.

\end{description}

\section{Autoregressive Models for Sequence Generation}

In this section we will focus on a particularly useful class of probabilistic models, which we call autoregressive generative models \citep[see e.\,g.\ ][]{theis2012mcgsm,larochelle2011nade,bengio2015scheduledsampling}. An autoregressive probabilistic model explicitly defines the joint distribution over a sequence of symbols $x_{1:N}$ recursively as follows:

\begin{equation}
	Q_{1:N}(x_{1:N}) = \prod_{n=1}^{N} Q_n(x_n\vert x_{1:n-1};\theta).
\end{equation}

We note that technically the above equation holds for all joint distributions $Q_{1:N}$, here we further assume that each of the component distributions $Q_n(x_n\vert x_{1:n-1};\theta).$ are tractable and easy to compute. Autoregressive models are considered relatively easy to train, as the model likelihood is typically tractable. This allows us to train even complicated deep models such as stacked LSTMs in the coherent and well understood framework of maximum likelihood estimation \citep{theis2012mcgsm, theis2015generative}.


\section{The symptoms\label{sec:thesymptoms}}

Despite the elegance of a closed-form maximum likelihood training, \citet{bengio2015scheduledsampling} have observed out that maximum likelihood training leads to undesirable behaviour when the models are used to generate samples from. In this section we review these \emph{symptoms}, and throughout this paper we will explore different strategies aimed at explaining and

Typically, when training an AR model, one minimises the log predictive likelihood of the $n$th symbol in each training sentence conditioned on all previous symbols in the sequence that we collectively call the prefix. This can be thought of as a special case of maximum likelihood learning, as the joint likelihood over all symbols in a sequence factorises into these conditionals via the chain rule of probabilities.

When using the trained model to generate sample sequences, we generate each new sequence symbol-by-symbol in a recursive fashion: Assuming we already generated a prefix of $n$ sybols, we feed that prefix into the conditional model, and ask it to output the predictive distribution for the $n+1$st character. The $n+1$st character is then sampled from this distribution and added to the prefix.

Crucially, at training time the RNN only sees prefixes from real training sequences. However, at generation time, it can generate a prefix that is never seen in the training data. Once an unlikely prefix is generated, the model typically has a hard time recovering from the mistake, and will start outputting a seemingly random string of symbols ending up with a sample that has poor perceptual quality and is very unlikely under the true sequence distribution $P$.


\section{Symptomatic treatment: Scheduled sampling}

In \citep{bengio2015scheduledsampling}, the authors stipulate that the cause of the observed poor behaviour is the disconnect between how the model is trained (it's always fed prefixes from real data) and how it's used (it's always fed synthetic prefixes generated by the model itself). To address this, the authors propose an alternative training strategy called scheduled sampling (SS). In scheduled sampling, the network is sometimes given its own synthetic data as prefix instead of a real prefix at training time. This, the authors argue, simulates the environment in which the model is used when generating samples from it.

More specifically, we turn each training sequence into modified training sequence in a recursive fashion using the following procedure:
\begin{itemize}
 \item for the $n$th symbol we draw from a Bernoulli distribution with parameter $\epsilon$ to decide whether we keep the original symbol or use one generated by the model
 \item if we decided to replace the symbol, we use the current model RNN to output the predictive distribution of the next symbol given the current prefix, and sample from this predictive distribution
 \item we add to the training loss the log predictive probability of the real $n$th symbol, given the prefix (the prefix at this point may already contain generated characters)
 \item depending on the coinflip above, the original or simulated character is added to the prefix and we continue with the recursion
\end{itemize}

The method is called scheduled sampling to describe the way the hyperparameter $\epsilon$ is annealed during training from an initial value of $\epsilon=1$ down to $\epsilon=0$. Here, we would like to understand the limiting behaviour of this training procedure, whether and why it is an appropriate way to address the shortcomings of maximum likelihood training.

\subsection{Scheduled sampling formulated as KL divergence minimisation}

To keep notation simple, let us consider the case of learning sequences of length 2, that is pairs of random symbols $x_1$ and $x_2$. Our aim is to formulate a closed form training objective that corresponds to scheduled sampling.

If $x_1$ is kept original - rather than replaced by a sample - the scheduled sampling objective in fact remains the same as maximum likelihood. We can understand maximum likelihood as minimising the following KL divergence\footnote{more precisely, maximum likelihood minimises the cross-entropy $KL[P\|Q] + H[P]$, where $H[P]$ is the differential entropy of training data.} between the true data distribution $P$ and our approximation $Q$:

\begin{align}
	D_{ML}[P\|Q] &= KL[P\|Q]\\
		&= KL[P_{x_1}\|Q_{x_1}] + \mathbb{E}_{z\sim P_{x_1}} KL[P_{x_2\vert x_1=z}\|Q_{x_2\vert x_1=z}] \label{eq:ML_divergence}
\end{align}

Here, $P_{x_1}$ and $Q_{x_1}$ denote marginal distributions of the first symbol $x_1$ under $P$ and $Q$ respectively, while $Q_{x_2\vert x_1=z}$ and $P_{x_2\vert x_1=z}$ denote the conditional distributions of the second symbol $x_2$ conditioned on the value of the first symbol $x_1$ being $z$.

The other case we need to consider is when $x_1$ is replaced by a sample from the model, in this case $Q_{x_1}$. The training objective can now be expressed as the following divergence:

\begin{align}
	D_{alternative}[P\|Q] &= KL[P_{x_1}\|Q_{x_1}] + \mathbb{E}_{y\sim P_{x_1}} \mathbb{E}_{z\sim Q_{x_1}} KL[P_{x_2\vert x_1=y}\|Q_{x_2\vert x_1=z}]\\
		&= KL[P_{x_1}\|Q_{x_1}] + \mathbb{E}_{z\sim Q_{x_1}} KL[P_{x_2}\|Q_{x_2\vert x_1=z}]\label{eq:fullSS_divergence}
\end{align}

Notice how in the second term the KL divergence is now measured from $P_{x_2}$ rather than the conditional, this is because the real value of the first symbol is never shown to the model, when it is asked to predict the second symbol $x_2$.

In scheduled sampling, we choose randomly between the above two cases, so the full SS objective can be described as a convex combination of $D_{ML}$ and $D_{alternative}$ above:

\begin{equation}
	D_{SS}[P\|Q] = KL[P_{x_1}\|Q_{x_1}] + \epsilon \mathbb{E}_{z\sim P_{x_1}} KL[P_{x_2\vert x_1=z}\|Q_{x_2\vert x_1=z}] + (1-\epsilon) \mathbb{E}_{z\sim Q_{x_1}} KL[P_{x_2}\|Q_{x_2\vert x_1=z}] \label{eq:SS_divergence}
\end{equation}

It is worth noting at this point that this divergence is an idealised form of the scheduled sampling. In the actual algorithm, expectations over $Q_{x_1}$ and $Bernoulli(\epsilon)$ would be implemented by sampling\footnote{The authors also propose taking argmax of each distribution instead of sampling, this case is harder to analyse but we think our general observations still hold.}. This divergence describes the method's limiting behaviour in the limit of infinite training data.

By rearranging terms we can further express the SS objective as the following KL divergence:

\begin{align}
	D_{SS}[P\|Q] &= KL[P_{x_1}\|Q_{x_1}] + \mathbb{E}_{z\sim P_{x_1}} KL\left[ \epsilon P_{x_1\vert x_1=z} + \frac{Q_{x_1}(z)}{Q_{x_1}(z)}P_{x_2} \middle\| Q_{x_2\vert x_1=z} \right] + C_{P,\epsilon}\\
		&= KL\left[P_{x_1}\left( \epsilon P_{x_1\vert x_1} + (1-\epsilon)\frac{Q_{x_1}P_{x_2}}{P_{x_1}}\right) \middle\| Q_{x_1,x_2} \right] + C_{P,\epsilon}
\end{align}

A very natural requirement for any divergence function used to assess goodness of fit in probabilistic models is that it is minimised when $Q=P$. In statistics, this property is referred to as strictly proper scoring rule estimation \citep{gneiting2007strictly}. Working with strictly proper divergences guarantees consistency, i.\,e.\ that the training procedure can ultimately recover the true $P$, assuming the model class is flexible enough and enough training data is provided. What the above analysis shows us is that scheduled sampling is not a consistent estimation strategy. As $\epsilon\rightarrow 0$, the divergence is globally minimised at the factorised distribution $Q=P_{x_1}P_{x_2}$, rather than at the correct joint distribution $P$. The model is still inconsistent when intermediate values $0<\epsilon<1$ are used, in this case the divergence has a global optimum that is somewhere between the true joint $P$ and the factorised distribution $P_{x_1}P_{x_2}$.

Based on this analysis we suggest that scheduled sampling works by pushling models towards a trivial solution of memorising distribution of symbols conditioned on their position in the sequence, rather than on the prefix of preceding symbols. In recurrent neural network (RNN) terminology, this would means that the optimal architecture under SS uses its hidden states merely to implement a simple counter, and learns to pay no attention whatsoever to the content of the sequence prefix. While this may indeed lead to models that are more likely to recover from mistakes, we believe it fails to address the limitations of maximum likelihood the authors initially set out to solve.

How could an inconsistent training procedure still achieve state-of-the-art performance in the image captioning challenge? There are multiple possible explanations to this. We speculate that the optimisation was not run until full convergence, and perhaps an improvement over the maximum likelihood solution was found as a coincidence due to the the interplay between early stopping, random restarts, the specific structure of the model class and the annealing schedule for $\epsilon$.


\section{The Diagnosis}

After discussing scheduled sampling, a method proposed to remedy the symptoms explained in section \ref{sec:thesymptoms}, we now seek a better explanation of why those symptoms exist in the first place. We will now leave the autoregressive model class, and consider probabilistic generative models in their full generality.

The symptoms outlined in Section \ref{sec:thesymptoms} can be attributed to a mismatch between the loss function used for training (likelihood) and the loss used for evaluating the model (the perceptual quality of samples produced by the model). To fix this problem we need a training objective that more closely matches the perceptual metric used for evaluation, and ideally one that allows for a consistent statistical estimation framework.

\subsubsection{A model of no-reference perceptual quality assessment}

When researchers evaluate their generative models for perceptual quality, they draw samples from it, then - for lack of a better word - \emph{eyeball} the samples. In visual information processing this is often referred to as no-reference perceptual quality assessment \citep[see e.\,g.\ ][]{wang2002noreference}. When using the model in an application like caption generation, we typically draw a sample from a conditional model $\mathbf{y}\vert\mathbf{x}\sim Q_{\mathbf{y}\vert\mathbf{x}}$, where $mathbf{x}$ represents the context of the query, and present it to a human observer. We would like each sample to pass a Turing test. We want the human observer to feel like $\mathbf{y}$ is a plausible naturally occurring response, within the context of the query $\mathbf{x}$.

In this section, we will propose that the KL divergence $KL[Q\| P]$ can be used as an idealised objective function to describe the no-reference perceptual quality assessment scenario. First of all, we make the assumption that the perceived quality of each sample is related to the \emph{surprisal} $- \log Q_{human}(x)$ under the human observers' subjective prior of stimuli $Q_{human}(\mathbf{x})$ CITE.
We further assume that the human observer has learnt an accurate model of the natural distribution of stimuli, thus, $Q_{human}(\mathbf{x}) = P(\mathbf{x})$. These two assumptions suggest that in order to optimise our chances in the Turing test scenario, we need to minimise the following cross-entropy or perplexity term:

\begin{equation}
	- \mathbb{E}_{\mathbf{x}\sim Q} \log P(x)\label{eq:perplexity}
\end{equation}

Note that this perplexity is the exact opposite average negative log likelihood $- \mathbb{E}_{\mathbf{x}\sim P} \log Q(x)$, with the role of $P$ and $Q$ changed.

However, the objective in Eqn.\ \ref{eq:perplexity} would be maximised by a model $Q$ that deterministically picks the most likely stimulus. To enforce diversity one can simultaneously try to maximise the entropy of $Q$. This leaves us with the following KL divergence to optimise:

\begin{equation}
	KL[Q\| P] = - \mathbb{E}_{\mathbf{x}\sim Q} \log P(x) + \mathbb{E}_{\mathbf{x}\sim Q} \log Q(x)
\end{equation}

It is known that $KL[Q\|P]$ is minimised when $P=Q$, therefore minimising it would correspond to a consistent estimation strategy. However, it is only well-defined when $P$ is positive and bounded in the full support of $Q$, which is not the case when $P$ is an empirical distribution of samples and $Q$ is a smooth probabilistic model. For this reason, $KL[Q\|P]$ is not viable as a practical training objective in statistical esimation. Still, we can use it as our idealised perceptual quality metric to motivate our choice of practical objective functions.

\subsubsection{How does this explain the symptoms?}

The differences in behaviour between $KL[Q\|P]$ and $KL[P\|Q]$ are well understood and exploited for example in the context of approximate Bayesian inference \citep{lacoste2011approximate,mackay2003information,minka2001EP}. The differences are most visible when model underspecification is present: imagine trying to model a multimodal $P$ with a simpler, unimodal model $Q$. Minimising $KL[P\|Q]$ corresponds to moment matching and has a tendency to find models $Q$ that cover all the modes of $P$, at the cost of placing probability mass where $P$ has none. Minimising $KL[Q\|P]$ in this case leads to a mode-seeking behaviour: the optimal $Q$ will typically concentrate around the largest mode of $P$, at the cost of completely ignoring smaller modes. These differences are illustrated visually in Figure \ref{fig:JSD_range}, panels B and D.

In the context of generative models this means that minimising $KL[P\|Q]$ often leads to models that overgeneralise, and sometimes produce samples that are very unlikely under $P$. This would explain why recurrent neural networks trained via maximum likelihood also have a tendency to produce completely unseen sequences. Minimising $KL[P\|Q]$ will aim to create a model that can generate all the behaviour that is observed in real data, at the cost of introducing behaviours that are never seen. By contrast, if we train a generative model by minimising $KL[Q\|P]$, the model will very conservatively try to avoid any behaviour that is unlikely under $P$. This comes at the cost of ignoring modes of $P$ completely, unless those additional modes can be modelled without introducing probability mass in regions where $P$ has none.

Once again, both $KL[P\|Q]$ and $KL[Q\|P]$ define consistent estimation strategies. They differ in the kind of errors they make under severe model misspecification particularly in high dimensions.

\section{Generalised Adversarial Training}

We theorised that $KL[Q\|P]$ may be a more meaningful training objective if our aim was to improve the perceptual quality of generative models, but it is impractical as an objective function.

Here we show that a generalised version of adversarial training \citep{goodfellow2014adversarial} can be used to approximate training based on $KL[Q\|P]$. Adversarial training can be described as minimising an approximation to the Jensen-Shannon divergence between $P$ and $Q$ \citep{goodfellow2014adversarial,theis2015generative}. The JS divergence between $P$ and $Q$ is defined by the following formula:

\begin{equation}
JSD[P\|Q] = JSD[P\|Q] = \frac{1}{2}KL\left[P\middle\|\frac{P+Q}{2}\right] + \frac{1}{2}KL\left[Q\middle\|\frac{P+Q}{2}\right]\label{eqn:JSD}
\end{equation}

\begin{figure}[t]
	\centering
	\input{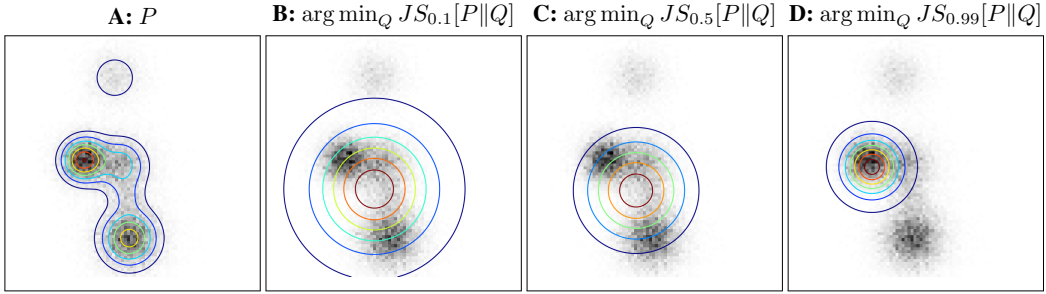}
	\caption{Illustrating the behaviour of the generalised JS divergence under model underspecification for a range of values of $\pi$. Data is drawn from a multivariate Gaussian distribution $P$ \textbf{(A)} and we aim approximate it by a single isotropic Gaussian \textbf{(B-D)}. Contours show level sets the approximating distribution, overlaid on top of the 2D histogram of observed data.
	For $\pi=0.1$, JS divergence minimisation behaves like maximum likelihood \textbf{(B)}, resulting in the characteristic moment matching behaviour. For $\pi=0.99$ \textbf{(D)}, the behaviour becomes more akin to the mode-seeking behaviour of minimising $KL[Q\|P]$. For the intermediate value of $\pi=0.5$ \textbf{(C)} we recover the standard JS divergence approximated by adversarial training. To produce this illustraiton we used software made available by \citet{theis2015generative}.}
	\label{fig:JSD_range}
\end{figure}

Unlike KL divergence, the JS divergence is symmetric in its arguments, and can be understood as being somewhere between $KL[Q\|P]$ and $KL[P\|Q]$ in terms of its behaviour. One can therefore hope that JSD would behave a bit more like $KL[Q\|P]$ and therefore ultimately tend to produce more realistic samples. Indeed, the behaviour of JSD minimisation under model misspecification is more similar to $KL[Q\|P]$ than $KL[P\|Q]$ as illustrated in Figure \ref{fig:JSD_range}. Empirically, methods built on adversarial training do tend to produce appealing samples \citep{goodfellow2014adversarial,denton2015lapgan}.

However, we can even formally show that JS divergence is indeed an interpolation between the two KL divergences in the following sense. Let us consider a more general definition of Jensen-Shannon divergence, parametrised by a non-trivial probability $0<\pi<1$:

\begin{equation}
JS_{\pi}[P\|Q] = \pi \cdot KL[P\|\pi P+(1-\pi)Q] + (1-\pi)KL[Q\|\pi P+(1-\pi)Q].\label{eqn:JSD_generalised}
\end{equation}

For any given value of $\pi$ this generalised Jensen-Shannon divergence is not symmetric in its arguments $P$ and $Q$ anymore, instead the following weaker notion of symmetry holds:

\begin{equation}
JS_{\pi}[P\|Q] = JS_{1-\pi}[Q\|P]\label{eq:weaksymmetry}
\end{equation}

It is easy to show that $JS_{\pi}$ divergence converges to $0$ in the limit of both $\pi \rightarrow 0$ and $\pi \rightarrow 1$. Crucially, it can be shown that the gradients with respect to $\pi$ at these two extremes recover $KL[Q\|P]$ and $KL[P\|Q]$, respectively. A proof of this property can be obtained by considering the Taylor-expansion $KL[Q\|Q+a] \approx a^THa$, where $H$ is the positive definite Hessian and substituting $a=\pi(P-Q)$ as follows:

\begin{align}
 \lim_{\pi\rightarrow0} \frac{JSD[P\|Q;\pi]}{\pi} &= \lim_{\pi\rightarrow0} \left\{ KL[P \| \pi P+(1-\pi)Q] + \frac{1-\pi)}{\pi}KL[Q \| \pi P+(1-\pi)Q] \right \}\\
 	&= KL[P\|Q] + \lim_{\pi\rightarrow0} \frac{1}{\pi} \pi^2(P-Q)^{T}H(P-Q)\\
 	&= KL[P\|Q]
\end{align}

Therefore, we can say that for infinitisemally small values of $\pi$, $JS\_{\pi}$ is approximately proportional to $KL[P\|Q]$:

\begin{equation}
 \frac{JS_{\pi}[P\|Q]}{\pi} \approx KL[P\|Q].
\end{equation}

And by symmetry in Eqn.\ \ref{eq:weaksymmetry} we also have that for small values of $\pi$

\begin{equation}
 \frac{JS_{1-\pi}[P\|Q]}{1-\pi} \approx  KL[Q\|P]
\end{equation}


This limiting behaviour also implies that for small values of $\pi$, $JS_{\pi}$ has the same optima as $KL[P\|Q]$. For values of $\pi$ close to $1$, $JS_{\pi}$ has the same optima as $KL[Q\|P]$. Thus, by minimising $JS_\pi$ divergence for a range of $\pi\in(0,1)$ allows us to interpolate between the behaviour of $KL[P\|Q]$ and $KL[Q\|P]$.

We note that $JS_{\pi}$ can also be approximated via adversarial training as described in \citep{goodfellow2014adversarial}. The practical meaning of the parameter $\pi$ is the ratio of labelled samples the adversarial disctiminator network receives from $Q$ and $P$ in the training procedure. $\pi=\frac{1}{2}$ implies that the adversarial network is faced with a balanced classification problem as it is the case in the original algorithm, for $\pi<\frac{1}{2}$ samples from the real distribution $P$ are overrepresented. Similarly, when $\pi>\frac{1}{2}$, the classification problem is biased towards $Q$. Thus, the generality of generative adversarial networks can be greatly improved by incorporating a minor change to the procedure. We note that this change may have adverse effects on the convergence properties of the algorithm, which we have not investigated. 

\section{Conclusions}

In this paper our goal was to understand which objective functions work and which ones don't in the context of generative models. Here we were only interested in models that are created for the purpose of drawing samples from, and we excluded other use-cases such as semi-supervised feature learning.

Our findings and recommendations can be summarised as follows:

\begin{enumerate}
 \item Maximum likelihood should not be used as the training objective if the end goal is to draw realistic samples from the model. Models trained via maximum likelhiood have a tendency to overgeneralise and generate unplausible samples.
 \item Scheduled sampling, designed to overcome the shortcomings of maximum likelihood, fails to address the fundamental problems, and we showed it is an inconsistent training strategy.
 \item We theorise that $KL[Q\|P]$ could be used as an idealised objective function to describe the no-reference perceptual quality assessment scenario, but it is impractical to use in practice.
 \item We propose the generalised Jensen-Shannon divergence as a promising, more tractable objective function that can effectively interpolate between maximum likelihood and $KL[Q\|P]$-minimisation.
 \item Our analysis suggests that adversarial training strategies are a the best choice for generative modelling, and we propose a more flexible algorithm based on our generalised JS divergence.
\end{enumerate}

While our analysis highlighted the merits of adversarial training, it should be noted that the method is still very new, and has serious practical limitations. Firstly, the generative adversarial network algorithm is based on sampling from the approximate model $Q$, which is highly inefficient in high dimensional spaces. This limits the applicability of these methods to low-dimensional problems, and increases sensitivity to hyperparameters. Secondly, it is unclear how to employ adversarial traning on discrete probabilistic models, where the sampling process cannot be described as a differentiable operation. To make adversarial training practically viable these limitations need to be addressed in future work.

\subsubsection*{Acknowledgments}

I would like to thank Lucas Theis and Hugo Larochelle for useful discussions. I would like to thank authors \citet*{theis2015generative} for kindly providing the source code for creating illustrations in Figure \ref{fig:JSD_range}.

\bibliography{generativemodels}
\bibliographystyle{iclr2016_conference}

\end{document}